\documentclass[letterpaper, 10 pt, conference]{ieeeconf}  

\IEEEoverridecommandlockouts                              

\usepackage{amsfonts}
\usepackage{graphicx}
\usepackage{todonotes}

\title{\LARGE \bf
Statistical shape representations for temporal registration of plant components in 3D
}

\author{Karoline Heiwolt$^{1}$, Cengiz Öztireli$^{2}$, Grzegorz Cielniak$^{1}$
\thanks{This work was supported by the Engineering and Physical Sciences Research Council [EP/S023917/1]}
\thanks{For the purpose of open access, the author(s) has applied a Creative Commons Attribution (CC BY) license to any Accepted Manuscript version arising.}
\thanks{$^{1}$KH and GC are with the Lincoln Centre for Autonomous Systems, University of Lincoln, Lincoln LN6 7TS, United Kingdom
        {\tt\small kheiwolt@lincoln.ac.uk}, {\tt\small gcielniak@lincoln.ac.uk}}%
\thanks{$^{2}$ CÖ is with the Department of Computer Science and Technology, University of Cambridge, Cambridge, United Kingdom
        {\tt\small aco41@cam.ac.uk}}%
}

\begin{document}

\maketitle
\thispagestyle{empty}
\pagestyle{empty}
\begin{abstract}
Plants are dynamic organisms and understanding temporal variations in vegetation is an essential problem for robots in the wild.
However, associating repeated 3D scans of plants 
across time is challenging. A key step in this process is re-identifying and tracking the same individual plant components over time. Previously, this has been achieved by comparing their global spatial or topological location. In this work, we demonstrate how using shape features improves temporal organ matching. We present a landmark-free shape compression algorithm, which allows for the extraction of 3D shape features of leaves, characterises leaf shape and curvature efficiently in few parameters, and makes the association of individual leaves in feature space possible.
The approach combines
3D contour extraction and further compression using Principal Component Analysis (\textit{PCA}) to produce a shape space encoding, which is entirely learned from data and retains information about 
edge contours and 3D curvature.
Our evaluation on temporal scan sequences of tomato plants 
shows, that incorporating shape features improves temporal leaf-matching. A combination of shape, location, and rotation information proves most informative for recognition of leaves over time and yields a true positive rate of 75\%, a 15\% improvement on 
sate-of-the-art methods. 
This is essential for robotic crop monitoring, which enables whole-of-lifecycle phenotyping.
\end{abstract}

\section{INTRODUCTION}
Monitoring plants across their whole life cycle is an important task in modern agriculture and agronomy. The ability to observe dynamic changes in plant morphology will allow farmers, breeders, and researchers to measure plant performance, understand genome-environment interactions, and identify indicators for plant stresses~\cite{das2019leveraging}. 
Robotic automation plays an increasingly important role in enabling high-throughput phenotyping directly in the field, by recording richer, more accurate information at an increased observation frequency~\cite{James2022Highthroughput}.
In robotics, the use of three-dimensional (\textit{3D}) sensors for constructing (e.g.~\cite{zollhofer2018state}) and maintaining (e.g.~\cite{krajnik2017fremen}) models of relatively rigid and structured environments over time are very active areas of research.
However, plants are non-rigid, growing organisms with high architectural complexity. Rapid changes in appearance, morphology, and topology, such as the emergence of new organs, are common. Additionally, external factors including lighting, temperature, and wind conditions cause both short-term and permanent deformations~\cite{tardieu2017plant}. 
  \begin{figure}[t!]
      \centering
      \hspace{-0.4cm}
      \includegraphics[width=0.5\textwidth]{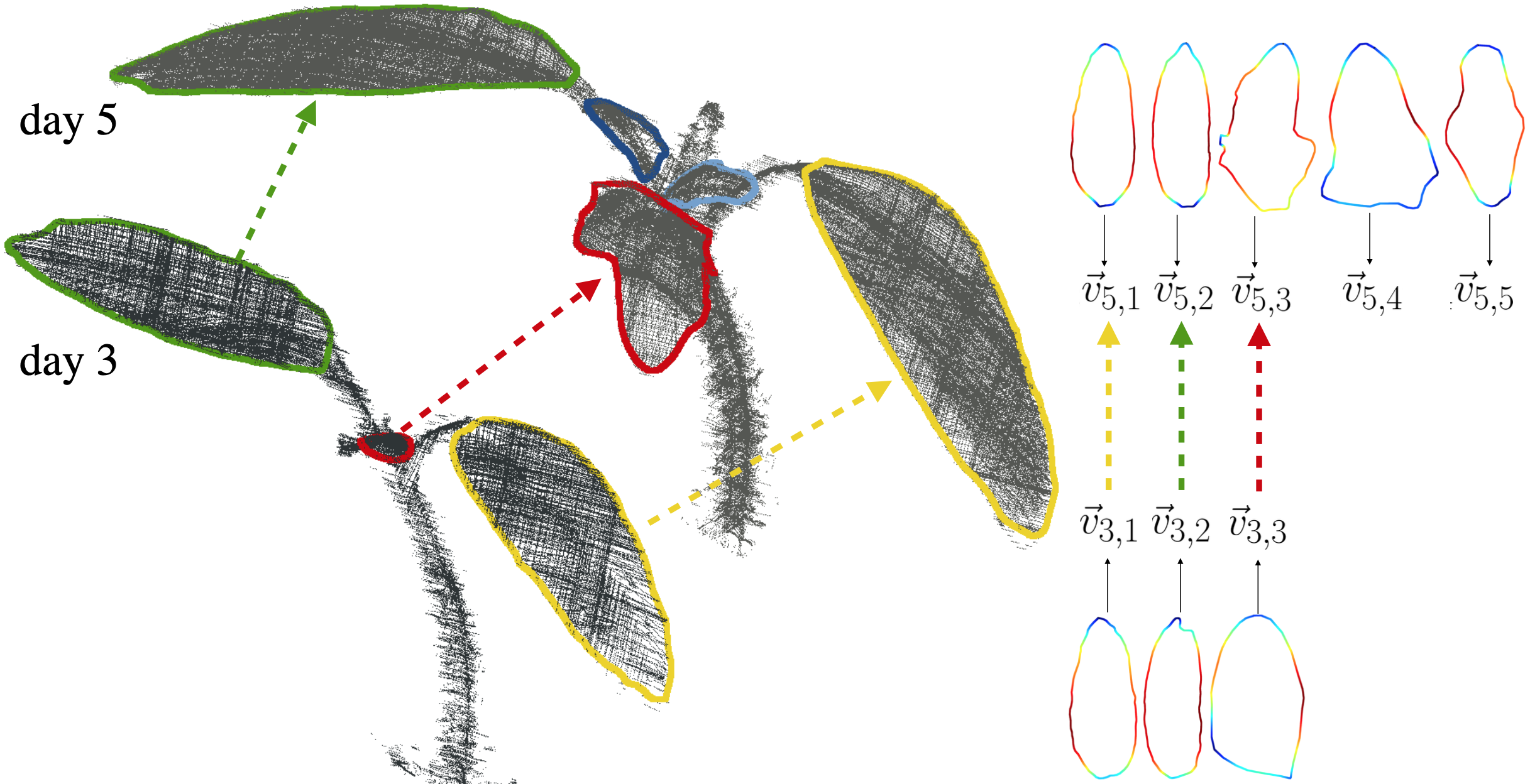}
      \caption{Left: 3D point clouds of the same tomato plant recorded two days apart, overlaid with 3D edge contours extracted using our proposed method. Right: 
      3D leaf outlines of both scans are compressed into feature vectors $v_{t,i}$ and matched in feature space.}
      \label{fig:overview}
   \end{figure}
A non-rigid spatio-temporal registration, which can locate correspondences between scans, while allowing for growth and appearance changes is needed. 

To cope with non-uniform growth patterns, segmenting plants into individual organs 
and tracking single components over time has proved helpful to simplify the registration problem~\cite{okura20223d, Magistri2020Segmentation, Zaganidis2018Integrating}. 
One essential step 
is the re-identification and tracking of the same organs in consecutive 3D scans. Existing approaches base the pairwise matching of unique leave instances 
on similarities in their global location in 3D space~\cite{Magistri2020Segmentation} or their relative topological location within the graph-like branching structure of the plant~\cite{Chebrolu2021registration, pan2021multi}.
Location proves informative across short time intervals and in controlled lab conditions with minimal external interference, such as wind. In the field, however, 
leaf location changes quickly and non-uniformly. Correspondence between leaves can not be guaranteed by a similar location. Especially if the time span between observations is increased, global location and topology are not reliable to robustly re-identify specific organ instances. Furthermore, topology is difficult to extract if the plant's branches can not fully be observed. In practice, due to heavy occlusions, the canopy can be observed more reliably than the hidden connectivity of leaves through a network of branches (e.g.~\cite{marks2022precise}).

In this paper, we propose to use 3D shape features 
to track unique plant components over time. 
We present a novel feature extraction pipeline to compute concise shape descriptors from 3D point clouds of plants and explore the use of these shape representations for temporal matching (Fig.~\ref{fig:overview}). We define the pairwise organ association task as a minimum bipartite graph matching problem, introduce a suitable feature space cost metric, and achieve robust association of leaves over time at 75\% accuracy without any manual intervention.
In summary, our contributions include:
\begin{itemize}
    \item A novel leaf shape compression pipeline, combining 3D contour extraction and subsequent PCA encoding.
    \item A cost metric designed to compare and associate plant components in shape feature space.
    \item Ablation and comparison studies demonstrating the applicability of this shape space representation to temporal leaf matching.
\end{itemize}

\section{RELATED WORK}
\subsection{Spatio-temporal registration of plant parts}
On top of a growing body of work focused on temporal registration on a whole-plot scale (e.g.~\cite{dong20174d}), recently, two approaches were introduced to address the spatio-temporal registration of individual plant point clouds:
In \cite{Chebrolu2021registration}, the authors perform point-wise binary semantic segmentation into leaves and stems, 
extract minimal skeleton representations of consecutive plant scans, and find
pairwise correspondences between nodes of temporally separated plant skeletons by describing the association as a Hidden Markov Model, with the hidden states representing the correspondences between nodes. The emission and the transition cost metrics exploit the skeleton's topology, i.e. the position of each node in the global graph-like organisation of the plant and their connectivity, as well as their location in Euclidean 3D space, and semantic class. 
Magistri and colleagues \cite{Magistri2020Segmentation} pair organs of the same semantic class by minimising the overall Euclidean distance between organ point clusters in two consecutive scans. Optimal assignment is computed using the Hungarian method~\cite{kuhn1955hungarian} on a cost matrix consisting of Euclidean distances between organ centroids.
In both cases, local leaf shape is not used explicitly. 

\subsection{Analysing leaf shape}
In several areas of plant sciences, edge contours are used as an informative proxy for 
leaf shape~\cite{backhaus2010leafprocessor}. On 2D images, leaf contours are found by manually marking predefined landmarks around the leaf perimeter (e.g.~\cite{biot2016multiscale}), or in an automated fashion, using tracing algorithms on binary images of single leaves~\cite{backhaus2010leafprocessor}. A standard approach to measuring variation and comparing shape in 2D is learning a statistical shape model for an object class from data. Most commonly, PCA is applied to compute the orthogonal vectors of greatest variance in a training data set and project the data onto those basis vectors. Individual examples of the class can then be described as a linear combination of shape feature components. 
In \cite{young1995morphometric}, the authors measure the distances between nine landmarks along the perimeter of leaves and perform PCA on those distances to detect shape differences in leaves 
grown in different environmental conditions. In~\cite{backhaus2010leafprocessor}, the authors perform PCA directly on outline coordinates to distinguish Arabidopsis leaf-shape mutants. \cite{victorino2019contour} introduces a method to perform PCA on Fourier Harmonics extracted from outline gradients, producing scale- and rotation-invariant shape descriptors to discover and distinguish interpretable shape categories in data sets of unlabelled 2D images of leaves, without relying on domain expertise. 
In each case, a low-dimensional vector space is found to concisely capture variability in shape within the given training set. 

However, 2D image projections inevitably fail to represent the characteristic curvature of leaves and can not capture the full growth process in 3D space. 
Related work on learning shape spaces from 3D data can be found in the fields of computer graphics, e.g. for recognition of faces~\cite{blanz1999morphable}, human skeleton tracking~\cite{allen2003space}, and detecting shape anomalies in medical imaging~\cite{rychlik2008applications}.
For PCA analysis all 3D shapes have to be aligned and described with the same corresponding topology, i.e. the same number of nodes with identical matrix connections. Objects are often registered by fitting a template mesh to each example and stacking the vertex coordinates into a single input vector~\cite{allen2003space}.
With our approach, we aim to extract a set of coordinates along the leaf edge, which captures both edge contour and 3D curvature. This representation is equivalent to fitting a template mesh, consisting of a single chain of vertices and edges of fixed length, to each leaf and thus establishing correspondence of coordinates.

\section{METHOD}
In the following, we first give a formal description of the organ assignment problem. 
Next, we present our proposed method, including 3D leaf outline extraction, shape space encoding via PCA, and computation of the feature space cost metric used to perform pairwise leaf matching. Finally, we define the baseline methods used for comparison. 

\subsection{Problem formulation}
The pairwise association of leaves between two subsequent scans can be described as a Minimum Bipartite Graph Matching problem.
A bipartite graph of leaves $G = (B, A, E)$ has two disjoint sets of vertices $B$ (before) and $A$ (after), such that each edge $(b_{i}, a_{j} ) \in E$ connects a vertex $b_{i} \in B$ with one $a_{j} \in A$. Each edge in graph $G$ has an associated cost $c_{ij} \in \mathbb{R}^{+}$.
A matching $M$ is a subset of edges $E$, such that no two edges in $M$ share a common vertex. The minimum cost bipartite graph matching is a matching for which the sum of costs across all edges is the minimum
\begin{equation}
c(M) = \sum_{(i,j) \in M}c_{ij}\,.
\end{equation}
This optimisation can be solved in polynomial time using the Hungarian method~\cite{kuhn1955hungarian}. An example is illustrated in the left panel of  Fig.~\ref{fig:BipartiteGraph}.

\subsection{Our approach}

\begin{figure}[t]
\centering
\vspace{0.2cm}
\includegraphics[width=0.45\textwidth]{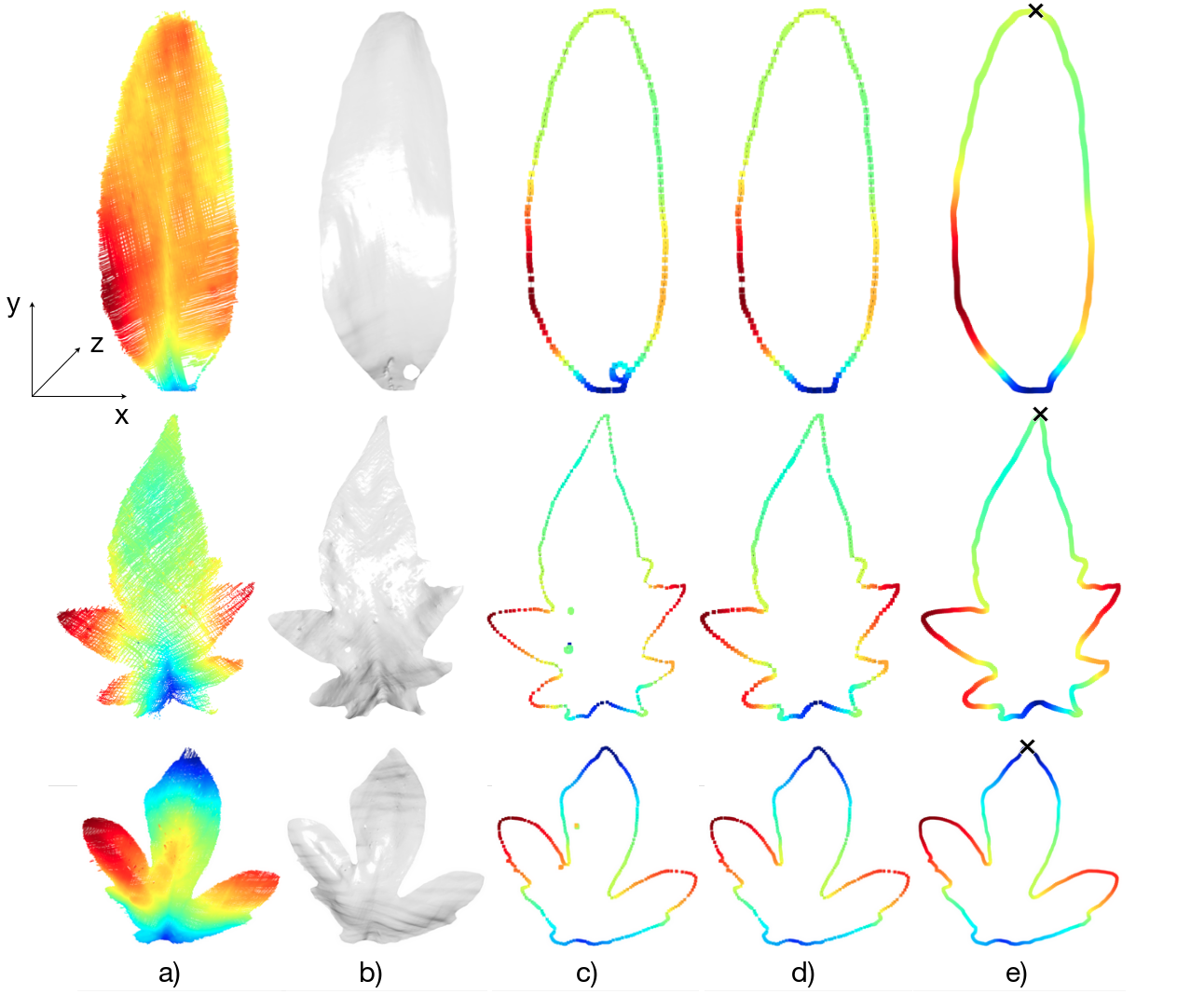}
\caption{The processing stages of boundary extraction, shown on three different leaf shapes. The columns show a) the aligned input point cloud, b) the triangulated mesh object, c) all mesh boundary loops, d) a single cleaned boundary loop, e) and a point set of $n=500$ outline coordinates, with the apex marked as \textbf{x}. Colours are added to illustrate depth.}
\label{fig:preprocessing}
\end{figure}

Our approach takes a time series of 3D plant point clouds with instance annotations for all leaves as input. To extract a concise summary of leaf shape, we first extract an outline representation of each leaf, consisting of a point set of $n=500$ coordinates sampled uniformly along its 3D edge contour. From a training set of outlines, we learn a statistical shape space, which allows the representation of individual leaves through a linear combination of concise shape descriptors. Pairwise assignment of leaves is performed using a cost metric based on distances between examples in shape feature space.

\subsubsection{Outline extraction}
The workflow comprises three steps: Determination of the main leaf axes, surface triangulation of the point cloud to obtain a concave triangle mesh, and extraction of a point set along the 3D mesh boundary.
Starting with the raw point clouds, all individual leaf point clusters are first aligned into a shared local coordinate frame, following the Leaf Axis Determination procedure in~\cite{ando2021robust}. The local right-hand coordinate frame for each leaf is defined by the leaf's principal axis, pointing in the direction of the largest variance in 3D space and oriented outward from the plant emergence point, and the axis perpendicular to the leaf surface (Fig.~\ref{fig:preprocessing}a).
Next, a triangle mesh object is computed from each point cloud via Delaunay Triangulation in 2.5D on the x-y-leaf plane (Fig.~\ref{fig:preprocessing}b).
A crucial parameter for mesh generation is the maximum permitted length $l_{max}$ for triangle edges, which represents a trade-off between maintaining fine detail in the contour along the edge of leaves and creating holes within the boundaries of leaves wherever the point sampling density is low. In favour of maintaining boundary detail, we set the maximum edge length low at $l_{max}=1$mm, and instead apply smoothing and hole closing operations to avoid holes in the mesh.
Before triangulation, we apply Laplacian smoothing exclusively in the z-dimension, i.e. perpendicular to the leaf surface, with a radius of $1.2$ and smoothing factor of $0.2$, for eight iterations. Following Delaunay Triangulation, a second step of Laplacian smoothing is applied in all three dimensions for 20 iterations. 
Finally, a hole filling operation is employed to close any holes with up to 10 edges. 

In the resulting mesh, a boundary loop is any chain of edges connecting those vertices, which are not enclosed within another complete loop of edges connecting its neighbouring vertices without connecting back to the vertex of interest (Fig.~\ref{fig:preprocessing}c). To filter out irrelevant closed loops around any remaining holes within the mesh, we only retain the single connected boundary loop with the largest spatial extent (Fig.~\ref{fig:preprocessing}d). Additionally, any leaf outlines for which at least one vertex is connected with two or more edges are excluded. This connectivity implies sub-loops within the contour and prevents uniform sampling along the edge in the next step. 

Finally, the boundary loop is sorted into clockwise order around the leaf in its local frame, starting at the maximum point along the main axis, which usually corresponds to the leaf apex, assuming proper leaf axis alignment. We then sample $n=500$ points around the outline, at uniform distances along the loop, starting at the apex (Fig.~\ref{fig:preprocessing}e). 
The coordinates are normalised by the total length of the outline to render the representation of shape scale invariant. Furthermore, the coordinate set can optionally be transformed back into global frame to re-introduce information about its global transform, specifically its location, relative rotation, and scale. The global frame origin is placed at the plant emergence point, which is considered a temporally consistent reference point to align all scans of the same plant over time. 
In summary, the outline representation retains information about the edge contour shape and 3D curvature of the leaf. If given in the leaf’s local coordinate frame with respect to the leaf apex at its origin, it is invariant to the scale, global location, and global orientation. Alternatively, we optionally translate, rotate, and scale outlines back to the original scan frame, retaining its original location, orientation, and scale relative to the plant emergence point. The representation achieves correspondence between points on the outlines of differently shaped leaves, which allows for the application of PCA.
 
\subsubsection{Limitations of outline extraction}
\begin{figure}[th]
    \centering
    \includegraphics[width=0.45 \textwidth]{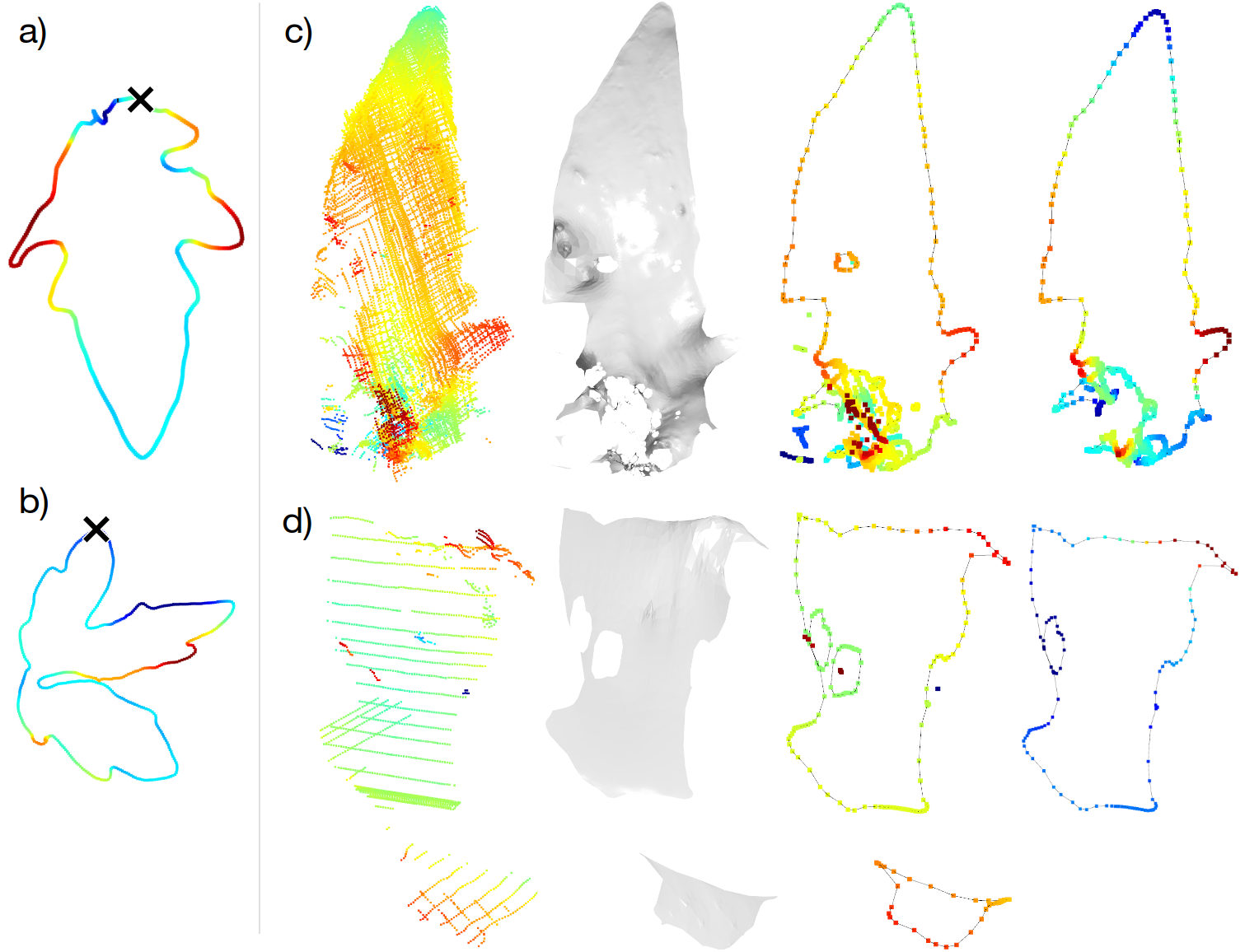}
    \caption{Error cases in the outline extraction procedure: In a) and b), axis alignment failed because of violated assumptions (13\% of leaves). In c) the sampling density is uneven across the leaf, and the leaf captured in d) is very small and thus the sampling density is low across the entire leaf.}
    \label{fig:errors}
\end{figure}
The assumption that leaves grow in an outward direction from the plant emergence point is violated for an average of 13\% of all leaves across any scans in the Pheno4D data set. These leaf point clouds are then oriented upside-down (Fig.~\ref{fig:errors}a) or less frequently sideways (Fig.~\ref{fig:errors}b) for the remainder of the outline extraction and as a consequence the apex point is misidentified. For our evaluation, we chose not to exclude these cases, but we found that shape space learning is improved when the correspondence between single leaves can be established with correctly aligned axes more consistently (see~\ref{sec:ablation}).
Furthermore, due to inadequacies of the sensing setup, the outline of leaves is sometimes ambiguous. Where the leaves are very small and/or the sensing density is low, it is difficult to extract complete mesh objects (see examples c) and d) in Fig.~\ref{fig:errors}). Future iterations of this work will explore alternative meshing algorithms so that more data can be considered for training. In our study, clean outlines could be extracted from 68\% of the original annotated data set.
The exclusion of those leaves emulates difficulties that are realistic to be expected in the field as well, considering that new leaves emerge and old leaves frequently die or are trimmed between time steps. A robust temporal matching algorithm is expected to cope with missing leaf instances.

\subsubsection{PCA shape encoding}
PCA computes a linear transformation which can be used to represent an input outline $o\in \mathbb{R}^{3n}$ with $n$ coordinates as a linear combination of $d$ shape parameters, which form a feature vector $ v \in \mathbb{R}^{d}$. The features are analogous to scalar weights $v_{k}$ in a $d$-dimensional subspace, spanned by the orthogonal vectors $V$, such that 
\begin{equation}
    o(v) = \bar{o} +  \sum_{k=1}^{d} v_{k}V_{k}\,,
\end{equation}
where $\bar{o}$ is the mean of all outlines in the training set.
The shape parameters are fit to training data and inversely, a shape vector can be projected back into 3D space. We apply $z$-score-normalisation to standardise feature scaling of the input outline vectors before fitting a PCA model to the training set.

\subsubsection{Cost metric}
The same feature scaling and PCA encoding are applied to the test set in all following experiments. To compute our cost metric, which is inversely proportional to the similarity between leaves in this multivariate shape feature space, we use the Mahalanobis distance metric. The Mahalanobis distance is a unitless and scale-invariant metric, measuring each feature's $z$-score along the corresponding axis, taking into account single features' distributions and potential correlations across the training data set. For a pair of PCA-encoded feature vectors $v_{t,i}$ and $v_{t+1,j}$ the cost is computed as
\begin{equation}
    c_{i,j} = \sqrt{(v_{t,i}-v_{t+1,j}) K^{-1}(v_{t,i}-v_{t+1,j})^{T}} \, ,
    \label{Mahalanobis_dist}
\end{equation} where 
$K$ is the covariance matrix of features computed across the training data set.

\subsection{Cost metrics based on transform}
In~\cite{Magistri2020Segmentation} the optimal pairwise matching is found using a cost based on the mean location of each leaf. 
Additionally, global rotation and scale information can be easily extracted from the leaf point cloud transform in the full plant's global reference frame. We represent any combination of location, rotation, and scale information as a one-dimensional feature vector. Location is given as the leaf centroid $m\in \mathbb{R}^{3}$, rotation as the three vectors describing the axes of the local leaf coordinate frame, 
given with respect to the global frame ${x}',{y}',{z}'\in \mathbb{R}^{3}$, and finally scale $s \in \mathbb{R}$, a scalar equivalent to the total length of the outline loop in the original coordinate frame.
Any combination of these transform descriptors are stacked in a one-dimensional vector $v = [m,{x}',{y}',{z}',s]$ with length between one and 13 elements. The cost $c$ is computed as the Mahalanobis distance, as defined in Eq.~\ref{Mahalanobis_dist}.

\section{EXPERIMENTAL EVALUATION}
In the following, we evaluate the quality of feature encoding and performance of our leaf-matching approach based on shape characteristics and various combinations of information about their global location, rotation, and size.

\subsection{Data set}
For testing, we use time-series 3D point cloud data of real tomato plants (Solanum Lycopersicum) taken from the Pheno4D set~\cite{schunck2021pheno4d}. It contains point clouds of 7 plants, captured once per day, across 20 days. 77 of the total 140 point clouds are annotated with consistent organ instance labels.
From those annotated scans, our outline extraction procedure yields 649 boundary instances, including leaves from all 7 unique plants, and spread across 11 time steps (on average two days apart). For the following experiments, we use leaf outlines from two plants for training (corresponding to 25\% of available data), and the remaining five for testing. This split ensures a sufficiently diverse training set of leaves from all growth stages.

\subsection{Performance metrics}
\label{sec:metrics}
\begin{figure}[t]
\centering
\vspace{0.2cm}
\includegraphics[width=0.44\textwidth]{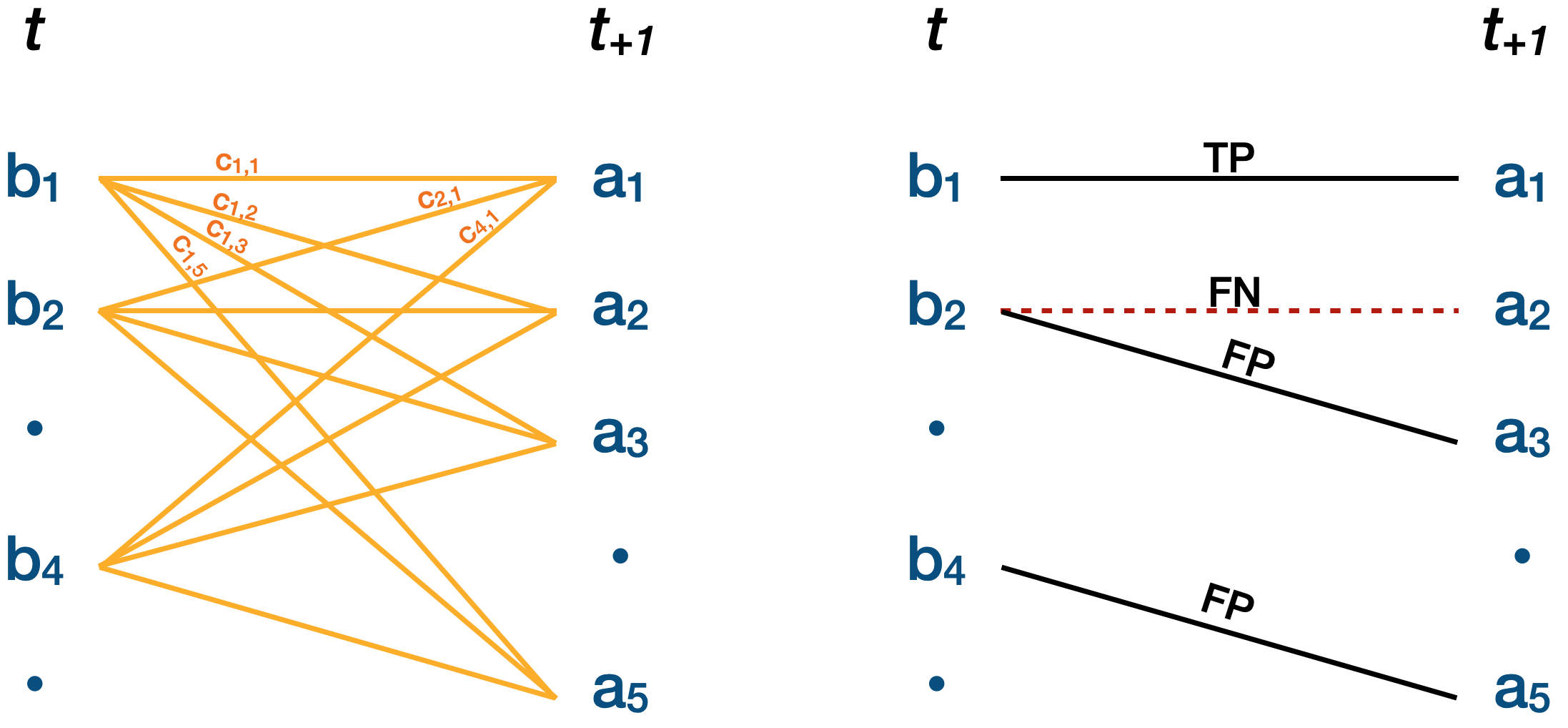}
\caption{\textbf{Left:} Example graph $G$ with all edges $E$ shown in yellow. Their associated costs $c_{i,j}$ are indicated in an exemplary way for the vertices b$_{1}$ and a$_{1}$. The vertices 3, 4, and 5 are only present in one of the two time steps $t$ or $t_{+1}$. \textbf{Right:} A predicted minimum cost matching (black lines) including examples of a true positive (TP) and false negative (FN), as well as an expected and an unexpected false positive (FP) pairing.}
\label{fig:BipartiteGraph}
\end{figure}

Typically, the Hungarian algorithm is applied to assign a set of $n$ jobs to a set of $m$ machines. If $n \neq m $, the algorithm finds a matching consisting of $min\{m,n\}$ edges, so that every vertex of the smaller set is covered.
In the case of plants, new components emerge and old components die or are trimmed frequently in between scans. These appearing and disappearing graph vertices are still included in the minimum cost matching (in Fig.~\ref{fig:BipartiteGraph} leaves 3 and 5 are missing at time $t$, leaf 4 is missing at time $t_{+1}$). As a result, the algorithm naturally predicts additional expected false positive (\textit{FP}) pairings, representing the best-fitting correspondences for any single leaves present in only one of the two scans. Thus, FP do not directly reflect the quality of the matching performance. However, we expect a suitable cost metric to yield a significantly higher associated cost for a false edge (\textit{F}), compared to true edges (\textit{T}), which include true positives (\textit{TP}) and false negatives (\textit{FN}, i.e. where a true pairing was not selected, see Fig.~\ref{fig:BipartiteGraph}b for an example). If the degree of separation between cost distributions for true corresponding edges compared to all possible false correspondence edges in $E$ is high, then a suitable cost metric could be used analogously to a confidence metric for suggested pairings, allowing for a threshold to be placed on the associated cost for predicted pairings in the future.
To reflect this requirement, we evaluate the sensitivity, or true positive rate, as
\begin{equation}
    \mathrm{sensitivity} = \frac{\mathrm{TP}}{\mathrm{TP} + \mathrm{FN}}
\end{equation} 
and the Distribution-Free Overlapping Index~\cite{pastore2019measuring} between observed cost distributions for true and false correspondence edges in $E$ across the test set as
\begin{equation}
    \eta = \frac{\sum_{i=1}^{b}min\{f^{T}_{i} \,\,,\, f^{F}_{i}\}}{min\{\mathrm{T} , \mathrm{F}\}}\,,
\end{equation}
where $f^{T}$ and $f^{F}$ are the observed discrete frequency densities of the cost associated with true and false edges respectively, discretised into the same $b=50$ bins. The measure is normalised to values between $0$ and $1$ by dividing by the number of edges in the smaller set. A value of $0$ indicates perfect separation, $1$ indicates perfect overlap of the cost distributions.

\subsection{Analysis of shape space encoding}
\begin{figure}[t]
    \centering
    \vspace{0.2cm}
    \includegraphics[width=0.46\textwidth]{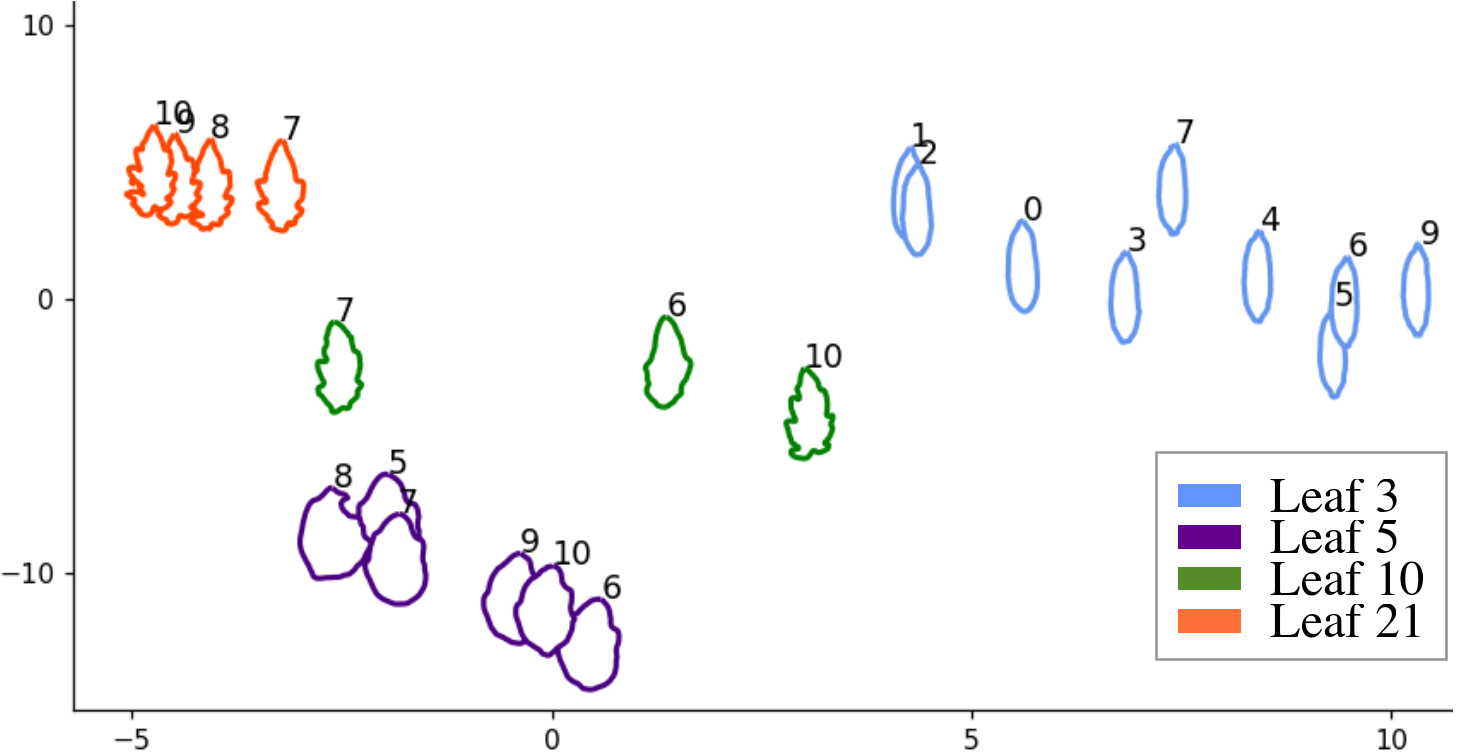}
    \caption{
    t-SNE visualisation of the learned shape representations for all instances of four example leaves. All outlines of the same colour are temporally separated instances of the same unique leaf. Number annotations indicate at which time step each outline was captured.}
    \label{fig:tsne}
\end{figure}

Fig.~\ref{fig:tsne} shows the t-distributed stochastic neighbour embedding (\textit{t-SNE}) for four individual leaves across time. This visualisation models the multivariate feature space distances between leaf instances in 2D and shows, how similar shapes cluster together. In the given example, differentiation between the four unique shown leaves in shape feature space is possible and thus the learned shape representation appears to capture useful information for associating leaves by shape similarities.
\begin{figure}[b]
    \centering
    \includegraphics[width=0.31\textwidth]{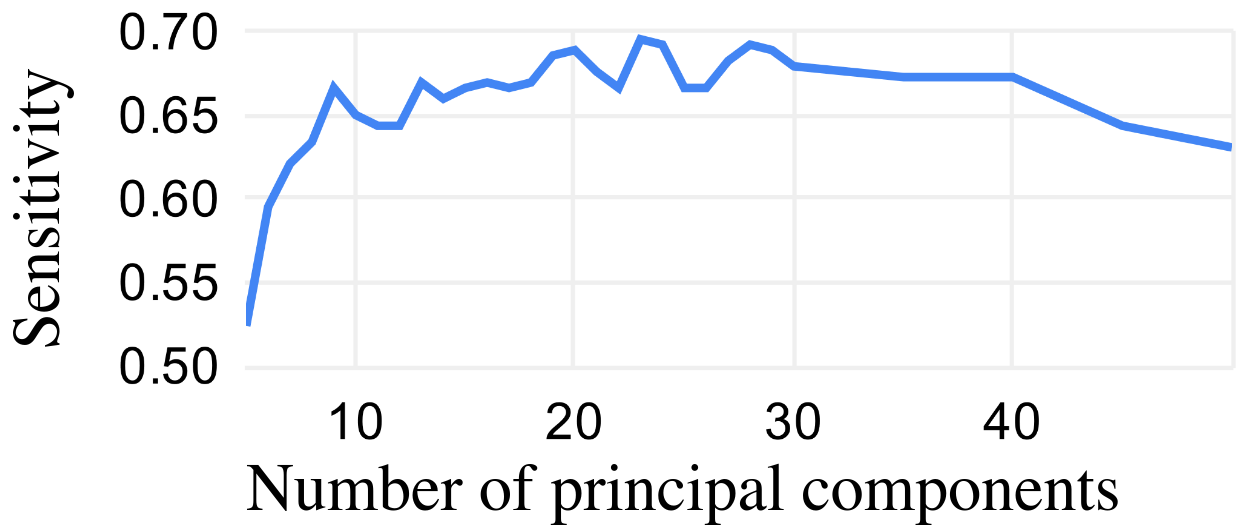}
    \caption{Sensitivity of outline matching procedure using shape, location, and rotation, as a function of the number of principal components considered.}
    \label{fig:components}
\end{figure}

\begin{figure*}
    \centering
    \vspace{0.2cm}
    \includegraphics[width=0.67\textwidth]{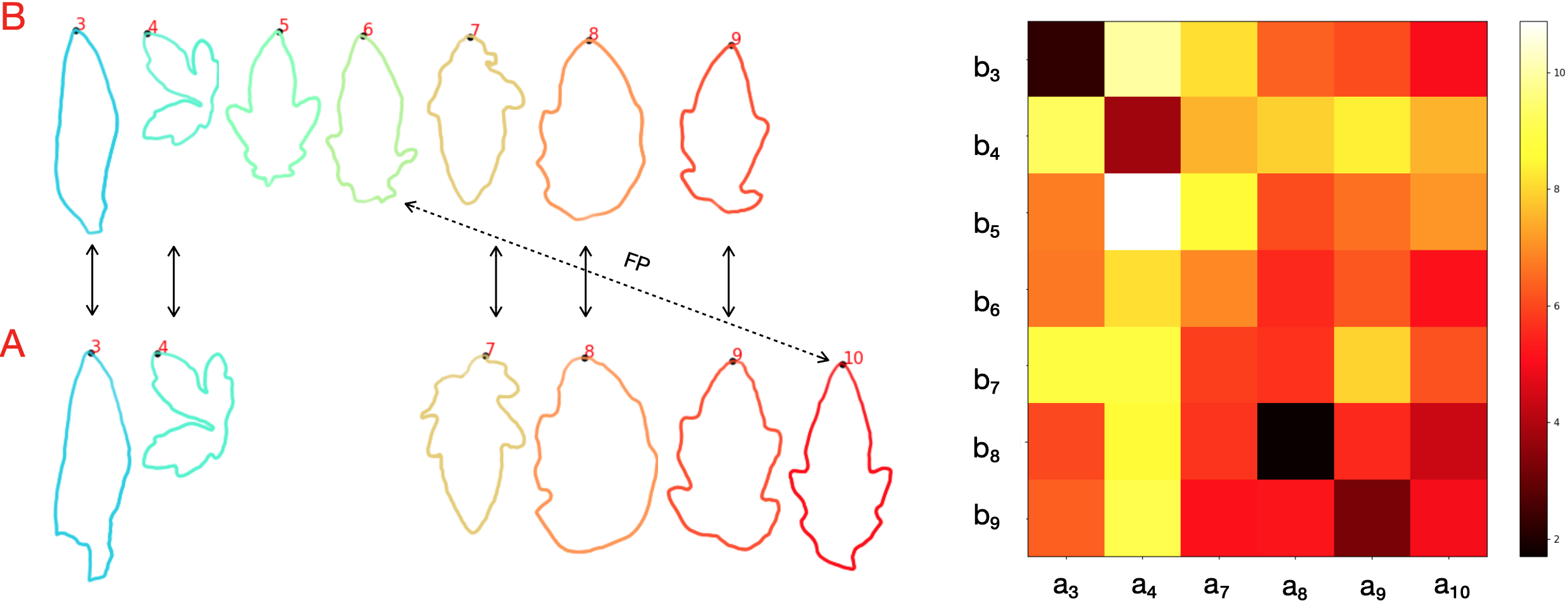}
    \caption{Example visualisation of the matching step. \textbf{Left:} All extracted leaf outlines observed at two consecutive time steps ($B$ - before and $A$ - after) as vertices of a bipartite graph. The arrows indicate edges, which are part of the minimum cost matching predicted by our method, including one natural false positive (FP) representing the best fit between new or disappearing leaves. \textbf{Right:} The confusion matrix of costs associated with each possible edge connecting elements of $B$ and $A$.}
    \label{fig:match_example}
\end{figure*}

Fig.~\ref{fig:match_example} shows an example of how the cost metric is then applied in practice to relate the organs of two consecutive 3D scans. The figure shows all extracted leaf outlines (left) and the confusion matrix of pairwise feature space distance or cost $c_{i,j}$ associated with each possible edge connecting vertices of $B$ and $A$. Like in Fig.~\ref{fig:BipartiteGraph}, black lines are used to show the matches predicted by our method. All true pairs are detected ($\mathrm{sensitivity} = 1$) and one additional expected false positive is predicted, representing the best-fitting pair of leaves present in only one of the two scans, as discussed in section~\ref{sec:metrics}. Note that leaf 7, which is misaligned and processed upside-down in both scans, is still assigned correctly. Across the full data set, we observed that this is generally the case, as long as both instances of the leaf are misaligned similarly.

Matching performance is further dependent on the amount of compression applied via PCA. Fig.~\ref{fig:components} illustrates the sensitivity of our outline matching procedure using shape only as a function of the number of principal components considered. We found that performance peaked in the range between 19 and 28 features. All following experiments were performed using the first 23 principal components for feature space compression.

\subsection{Ablation study}
\label{sec:ablation}
Table~\ref{tab:ablation} shows the sensitivity and overlapping index observed for the test set of 489 leaves taken from 5 plants across 11 time steps. 
The combination of shape, location, and orientation information proves most informative, achieving an overall true positive rate of 75.4\%.
While the difference in means between observed discrete frequency densities of cost associated with true and false edges respectively proved very highly significant for all ablations in independent samples t-tests (all $p<0.001$), the proportion of overlap varies broadly.
The biggest separation between observed cost distributions is achieved by our model using shape, location, and orientation information.
We further observed that adding scale information appears to hinder the successful learning of informative shape representations. By normalising outline length and thus removing uniform scale differences, the shape model learned through PCA only considers shape differences and not size differences. Reintroducing scale information, however, might interfere with the effort to establish correspondence between outline points and impair the model's ability to sufficiently capture shape. 
Additional to the results shown in table~\ref{tab:ablation}, we found that removing any misaligned leaves such as depicted in~\ref{fig:errors}a) and b) improved our best model by a further 10 percentage point margin to 85.5\% sensitivity and an overlapping index of 0.4298. However, this was achieved by visual examination and manual exclusion of individual leaves and an automatic filter does not currently exist in our implementation.

\begin{table}[t]
\caption{Ablation study}
\vspace{-0.1cm}
\begin{tabular}{cccc|ll}
\multicolumn{4}{c}{Ablations}  & &\multicolumn{1}{c}{}   \\
shape & location & rotation & scale & \multicolumn{1}{c}{sensitivity}  & \multicolumn{1}{c}{$\eta$}\\ \hline
\checkmark  &  &  &  & 0.6958 & 0.658\\
\checkmark  & \checkmark &  &  & 0.7087 & 0.6084\\
\checkmark   &          & \checkmark        &       & 0.6958 & 0.67\\
\checkmark   &          &          & \checkmark     & 0.6343 & 0.9935\\
\checkmark   & \checkmark        & \checkmark        &       & \textbf{0.7540} & \textbf{0.5275}\\
\checkmark   & \checkmark        &          & \checkmark     & 0.6958 & 0.8673\\
\checkmark   &          & \checkmark        & \checkmark     & 0.5955 & 0.9935\\
\checkmark   & \checkmark        & \checkmark        & \checkmark     & 0.7152 & 0.8382
\end{tabular}
\label{tab:ablation}
\end{table}


\subsection{Comparison study}
This experiment is designed to support the claim that shape features are a useful similarity measure for spatio-temporal tracking of unique plant components between temporally separated point clouds. We compare our feature space cost metric to alternative cost metrics based on transform information in Table~\ref{tab:comparison}. 
In our tests, for tomato plants scanned on average two days apart, our best method constitutes a 15\% sensitivity increase from alternative methods found in literature and 9\% improvement from the best alternative exploiting global leaf transform information. Note also that the shape-only ablation of our method still outperforms any method without shape information. In this experiment, again all methods produced cost distribution for true and false graph edges, which differed significantly in their mean, but the overlap between the cost distributions could be reduced notably from 0.9288 when considering location only to 0.5275 with our approach. 

\begin{table}[t]
\caption{Comparison study}
\vspace{-0.1cm}
\begin{tabular}{cccc|lll}\multicolumn{4}{c}{Method}  & &\multicolumn{1}{c}{}   \\
shape & location & rotation & scale & \multicolumn{1}{c}{sensitivity}  & \multicolumn{1}{c}{$\eta$}&\\ \hline
\checkmark   & \checkmark        & \checkmark        &       & \textbf{0.7540} & \textbf{0.5275} & (ours)\\
 & \checkmark   &  &  & 0.6051 & 0.9288 &~\cite{Magistri2020Segmentation} \\
 &  & \checkmark &  & 0.4822 &  0.9676\\
 &  &  & \checkmark & 0.2654 & 1\\
   &  \checkmark & \checkmark &  & 0.6117 & 0.8382\\  
  & \checkmark &  & \checkmark & 0.6634 & 0.9029\\
 &  & \checkmark & \checkmark &  0.5307 & 0.8997\\ 
    &  \checkmark & \checkmark & \checkmark  & 0.6246 & 0.8188\\  
\end{tabular}
\label{tab:comparison}
\end{table}

\section{CONCLUSION}
We propose a novel workflow to track individual leaves in sequences of 3D scans of plants by computing correspondences in shape feature space. We confirm our hypothesis that shape information is more informative for the task than location or additional leaf transform information. Our experiments suggest, that our approach outperforms the state of the art and a combination of shape, location, and rotation proved most informative in an ablation study. 
Future work will include improvements to the boundary extraction pipeline, including exploring alternative meshing algorithms. Furthermore, PCA is a common but limited statistical tool to learn shape spaces, this could be extended by employing feature learning such as through 
autoencoder networks. On top of that, alternative boundary representations might be employed to ensure full rotational invariance. Eventually, we aim to use the data association achieved with this method to establish time-series prediction models for plant growth.
Ultimately, robust temporal registration of plant components in 3D is an important underpinning technology needed to enable sustained crop monitoring with a large potential positive impact on the integration of agricultural robots into a sustainable global food chain, as well as applications in vegetation mapping for robots in natural environments.

\addtolength{\textheight}{-10.5cm}   





\typeout{}
\bibliographystyle{IEEEtran}
\bibliography{root.bib}


\end{document}